\documentclass[runningheads]{llncs}
\usepackage[T1]{fontenc}
\usepackage{graphicx}
\usepackage{enumitem}

\usepackage{algorithm}
\usepackage{algorithmic}
\usepackage{booktabs}
\usepackage[table]{xcolor}

\usepackage[most]{tcolorbox}
\newtcolorbox{myprompt}[1][]{
    colback=lightgray!5,
    colframe=cyan!75!black,
    fonttitle=\bfseries,
    title=#1,
    sharp corners,
    boxrule=0.5pt,
    boxsep=5pt,
    left=5pt,
    right=5pt,
    before skip=10pt,
    after skip=10pt,
    breakable
}

\usepackage{tikz}
\usetikzlibrary{shapes, arrows, positioning}

\usepackage{newfloat}
\usepackage{listings}

\title{SNAP: Semantic Stories for Next Activity Prediction}

\begin{document}

\title{SNAP: Semantic Stories for Next Activity Prediction}

\author{Alon Oved \and Segev Shlomov \and Sergey Zeltyn \and
Nir Mashkif \and Avi Yaeli }
\authorrunning{A. Oved et al.}
%
\institute{IBM Research - Israel \\
\email{alon.oved@ibm.com, segev.shlomov1@ibm.com, sergeyz@il.ibm.com,\\ nirm@il.ibm.com, aviy@il.ibm.com}}

\maketitle              

\begin{abstract}
Predicting the next activity in an ongoing process is one of the most common classification tasks in the business process management (BPM) domain. 
It allows businesses to optimize resource allocation, enhance operational efficiency, and aid both in risk mitigation and strategic decision-making. 
This provides a competitive edge in the rapidly evolving confluence of BPM and AI.
Existing state-of-the-art AI models for business process prediction do not fully capitalize on available semantic information within process event logs. As current advanced AI-BPM systems provide semantically richer textual data, the need for new adequate models grows. 
To address this gap, we propose the novel SNAP algorithm that leverages language foundation models by constructing narratives and semantic contextual stories from the process’s historical event logs and uses them for the next activity prediction. 
We compared the SNAP method with eleven state-of-the-art models on six benchmark datasets and show that SNAP significantly outperforms them on datasets with high levels of semantic content.

\end{abstract}
\section{Introduction}

\label{sec:introduction}

Predictive business process monitoring (PBPM) is a valuable tool for organizations. It allows them to anticipate potential risks and bottlenecks within their processes and take proactive measures to mitigate them. A PBPM technique aims to predict the future state of a process execution by using predictive models constructed from historical process event data logs. 
Currently, machine learning (ML) algorithms \cite{ramamaneiro2021benchmark} and process mining techniques \cite{van2011time} are the most prevalent approaches for developing PBPM models. These predictive models are usually designed for such tasks as next activity prediction, trace suffix prediction, process outcome prediction, and remaining time prediction \cite{neu2021prediction}. 

A frequently researched topic in PBPM \cite{neu2021prediction}, next activity prediction (NAP), aims to forecast the next step in a running process instance. NAP is mainly used in the following four applications: (1) next-best action recommendation - advise the next-best activity based on selected key performance indicators (KPIs) \cite{weinzierl2020next}; (2) anomaly detection - identify a deviating process instance compared with the distribution of the next activity predictions \cite{nolle2022binet}; (3) resource allocation - proactively assign resources based on activity prediction \cite{park2019prediction}; and (4) preemptive guidance - monitor the most plausible activities to flag inefficiencies, risks, and mistakes \cite{di2022predictive,ly2015compliance}.

Robotic process automation (RPA) \cite{aalst2018rpa} refers to the use of software to perform repetitive tasks that are typically carried out by humans. Such systems could be significantly aided by using PBPM tools such as NAP \cite{chakraborti2020robotic}. 
In the world of conversational RPA \cite{rizk2020conversational}, chatbot UI includes a text input control for user utterances which trigger automations, a messaging-like area to present the chatbot response, and the history of the conversation. 
These systems produce traces of the user-bot interactions. 
Each trace consists of identifiers, the selected skills (i.e., activities/automations), timestamps, the corresponding utterances, and other involved resources. 
In such systems, the NAP model has the potential to serve as a platform for recommending the next best automation skill \cite{yaeli2022recommending}. Such recommendations can be highly valuable, especially in instances where a vast array of automation capabilities are made available for user interactions. 
We hypothesize that incorporating the semantic context information carried in these attributes and utterances may enhance the NAP model's performance.

We observe that almost all of the existing solutions for next activity prediction utilize only the sequence of activities as input to generate a classification model. Rarely are the additional numerical and categorical attributes taken into account within such a framework for predictions. The main reason for the trend to rely mainly on the activity sequence, rather than also integrating natural language-based utterances and contextual inputs, is that it reduces  the complexities involved. 

To close this gap, we propose the novel Semantic stories for the Next Activity Prediction (SNAP) model.
The main idea is to use the richness of language to build a coherent story which can capture more details of the business process. SNAP's input is the business process dataset, which typically includes many features, where some of them are numerical, categorical, free text, or have little or no semantic meaning. The algorithm's output is a trained next-activity prediction model.

The main steps of the algorithm are as follows: (1) design the list of features to use in the stories; (2) prompt an LLM to acquire a story template (i.e., narrative) of the business process; (3) transform the event log into the semantic stories; (4) using these contextual stories, fine-tune a pre-trained language foundation model (LFM) for NAP as a classification task.

We conducted an extensive experimental study to analyze the proposed SNAP algorithm using three pre-trained LFMs: BERT, DeBERTa, and GPT-3. The empirical results demonstrate that SNAP can improve the next activity prediction performance for various BPM datasets with respect to Deep Learning (DL) benchmarks. We also show an outstanding increase in results for a dataset in the conversational Robotic Process Automation (RPA) domain, where richer semantic information exists.
To conclude, this paper's main contributions include:
\begin{itemize}[topsep=0pt]
    \item SNAP's core idea of using the richness of language for transforming processes into semantic stories
    \item SNAP's state-of-the-art results for the next activity prediction task using the most popular PBPM datasets, including a conversational RPA domain.
\end{itemize}

The rest of the paper is organized as follows. Section 2 introduces relevant work. Section 3 outlines a formal definition of the problem and details the proposed SNAP algorithm. 
In Section 4, we start with the description of the explored datasets.
Next, we present the results of the experiments and an analysis of SNAP performance compared with the state-of-the-art models. Section 5 includes a discussion on results and limitations of our approach. Finally, we propose future research directions. 

\section{Related work}

\label{sec:related_work}

Next activity prediction (NAP) and other PBPM tasks have a research history of several decades and numerous business applications. Initially, PBPM research used classical machine learning (ML) methods; Márquez-Chamorro et al.\cite{marquez2017survey} survey these techniques. 
Typically, ML methods, including SVM, Random Forests, and others, easily incorporate numeric and categorical attributes, but do not work well with sequential data and struggle with free-text and semantic information. Their performance is also typically inferior to deep learning (DL) methods.

Advanced DL models, such as LSTM, provided the best results for next activity prediction before the advent of transformers. \cite{neu2021prediction} contains a detailed review of this domain and discusses other aspects of the prediction pipeline, such as feature engineering and encoding. These methods are well suited for relatively short sequential data, but still have trouble incorporating free-text features, semantics, and long-term dependencies.

Currently, the transformer-based architectures \cite{vaswani2017attention} have become the most prevalent approach to natural language processing (NLP). These methods, boosted with large-scale corpora of text data, gave rise to language foundation models. Bi-directional approaches such as BERT \cite{devlin2018bert} and  DeBERTA \cite{he2020deberta}, as well as autoregressive approaches such as the GPT series \cite{dale2021gpt}, have outperformed the previous generation of models in a wide range of natural language processing (NLP) tasks such as sentiment analysis, text classification, machine translation, question answering, and more. 

Applying transformer-based models to the PBPM domain and NAP task raises several technical challenges. In one of the first papers on the topic, \cite{philipp2020attention} uses sequences of activities from event logs, where activity types are encoded to integers. An attempt to use BERT for NAP is shown in \cite{chen2022multi}. Their method performs a pre-training phase with a masked activity model (MAM) task and adds a classification head for the fine-tune phase. Similarly to \cite{philipp2020attention}, the input for the model is only the sequence of activities from the process trace. 
Recent research in \cite{ni2023transformer} uses hierarchical transformer and several methods for attribute encoding. However, textual encoding is yet to be solved. 

An additional transformer-based model, called TaBERT \cite{yin2020tabert},  uses tabular data as input, the same form of event logs data used for the NAP task. However, the TaBERT model cannot be applied for NAP as its objective is to solve query understanding in semantic parsing cases. 

In summary, most state-of-the-art models for NAP which utilize transformer-based architectures struggle to use semantic and textual content present in many advanced process logs, such as conversational RPA systems. In addition, they do not include many attributes in the input data due to state-space explosion. 

Comparing the different DL prediction techniques presents technical difficulties, as each approach has its own prediction settings and data cleaning procedure. For example, a prediction model can be initiated from the first, second, or later process steps. The set of activities used in the prediction task can either include or exclude a "dummy activity" for process end. 

\cite{ramamaneiro2021benchmark} addresses this core challenge by considering 12 open-access BPM datasets, running and comparing multiple publicly available implementations for several prediction tasks using the same settings. Specifically, nine DL algorithms for NAP are compared. \cite{hinnka2020rnn} uses a gated recurrent unit (GRU) DL technique that provides the highest accuracy for five datasets, while LSTM-based \cite{tax2017lstm} is the best model for four datasets. \cite{theis2019decay} uses a deep feed-forward neural network (NN) approach that wins in two datasets. The last valuable approach is using convolutional NN
\cite{Pasquadibisceglie2019cnn}; it provides the best accuracy for one dataset. 
In \cite{ramamaneiro2023graph}, the authors of \cite{ramamaneiro2021benchmark} provided their own approach for next activity prediction using graph convolutional networks. Their results improved SOTA for all except two datasets. 

Recent advances in goal-driven chatbots and RPAs \cite{chakraborti2020robotic} gave rise to a new class of business processes. \cite{rizk2020conversational} describes an approach to design and orchestration of such systems.  RPA systems extensively use chatbots and automation tools to interact with human process participants and automate business tasks.  These applications include user utterances and bot responses in event logs, resulting in new types of activities. A conversation session with a chatbot constitutes a sub-process. The combination of a user utterance and a chatbot response can be considered a basic activity. As far as we know, this type of real-world BPM dataset is still unavailable for public access. \cite{zeltyn2022mip} partially closes this gap by providing a synthetic dataset based on an actual use-case in the Human Resources domain. 

Our research aims to investigate an algorithm with which semantic context in process event logs can be leveraged to predict the next activity in the PBPM domain, utilizing the novel concept of process semantic stories. 

\section{The model}

\label{sec:methodology}

\subsection{Problem description}
The building blocks of business process management (BPM) include activity, case ID, traces, and others. These elements form the basic structure of a BPM system and are essential for defining, modeling, and executing business processes.
\textit{Activities} are units of work executed within a process. Activities can be sequential or parallel and can be defined using various attributes such as roles, resources, and dependencies. These attributes play a significant role in the process and may contain free text values.
The \textit{case ID} is a unique identifier assigned to each instance of a process. It allows organizations to track and manage individual cases and provides a way to aggregate data and analyze process performance. 

\textbf{Event logs and traces} - 
A trace consists of a finite sequence of events executed in a case, each defined by an activity, timestamp, and several attributes. An event log is the set of all traces executed in a process. 

\textbf{Next activity prediction} - An event with $L$ different attributes is defined as follows: $e= \left(a,t,(d_1,v_1),...,(d_L,v_L) \right)$, where $a$ is the activity, $t$ is its timestamp, $d_i$ is an attribute name, and $v_i$ is the corresponding attribute value.
Given a trace prefix such that $p_k=\langle e_1, e_2,..., e_k \rangle$, we aim to predict the next event's activity, $a_{k+1}$, from the trace prefix (i.e., $f(p_k)=a_{k+1}$).

%
\subsection{SNAP}
\label{subsec:snap}
%

The main idea of the SNAP model is based on the concept of semantic stories. 
Through narratives and special structure, stories include semantics and meaning, making them a universal elements of the human experience. 
In the context of processes, stories can help describe and structure a process and add meaning to the process data. Thus, to utilize all the available information in the process event log, we construct a story that captures the semantic information of each prefix trace. It provides a unified textual representation of the traces that allow us to fine-tune LFMs and achieve state-of-the-art results.
The general steps of SNAP are presented in Algorithm \ref{alg:algorithm}.

\begin{algorithm}[h]
\caption{SNAP}
\label{alg:algorithm}
\textbf{Input}: Event log $(E)$\\
\textbf{Output}: A story-based classification model\\
\begin{algorithmic}[1] 
\STATE Build the list of features for the story narrative generation by performing feature selection and temporal features design. 
\STATE Generate a story narrative from the features list, process domain, and feature values by prompting an LLM.
\STATE Transform each prefix trace in the event log into a coherent semantic story via the story narrative.
\STATE Fine-tune a Language Foundation Model where\\
    \hspace{0.5cm} Sample = story\\
    \hspace{0.5cm} Label = next activity\\
\end{algorithmic}
\end{algorithm}

\textbf{List of attributes} - 
The original event logs consist of $L$ different attributes. In many cases, only a few features should be part of the semantic story. There are two main reasons for that assertion. First, an LFM is limited in its input size (e.g., the 512 token limit in BERT), and second, we argue and empirically test that not all attributes are statistically significant for the classification task. We suggest using an XGBoost classification model for feature selection and set a threshold on the attributes' importance. 

Temporal features are computed from trace timestamps. For example, the time from the case start and the time from the previous activity is computed and added to the features list as a new temporal feature. 

\begin{figure}[h!]
    \centering
    \includegraphics[width=\textwidth]{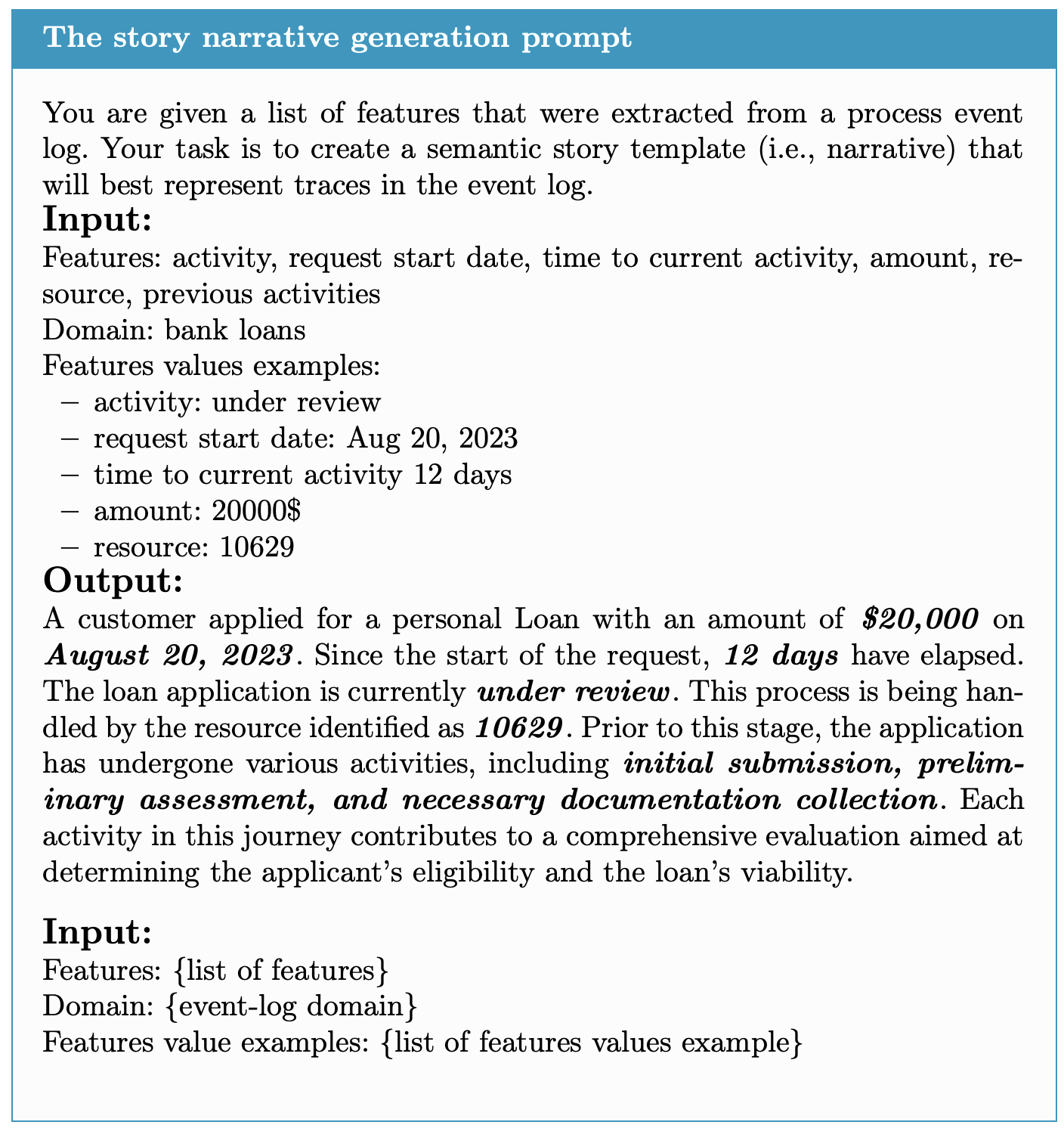}
    \caption{The LLM prompt used to create a story narrative. It includes a story output example and its corresponding input features.}
    \label{fig:prompt-example}
\end{figure}

\textbf{Story template construction} - 
To transform each prefix trace into a story we first construct a story template. To do so, we suggest utilizing a generative Large language model (LLM) (e.g., Llama 2 \cite{llama2}), as these types of models excel at such tasks.
Given the list of features from the previous step, SNAP uses an LLM prompt to generate a template story. We offer several few-shot examples and recommend using at least one in the prompt. Clearly, these samples can be modified to a specific required domain. Figure \ref{fig:prompt-example} illustrates an example of the one-shot prompt we used. 

The output generated by the LLM serves as a basis of the story template. Since this step is performed by an LLM, we advise to review its result and verify its coherence. In some instances, we might prefer a shorter story or the removal of unwanted information. In such cases, we can then manually edit the story template.

\textbf{Semantic story transformation} - 
Every prefix trace in the event log is transformed into a semantic story based on the process information, the case data, the attributes relevant to the current turn, and the previous activities executed in the case. 
For example, the prefix trace of a loan application could contain the following features (or values): 
Activity (Register Application), Turn (6), Time from case start (12 days), Support line (Customer Support), Latest impact (Moderate), Owner country (USA), Involved ST function div (Finance Division), Involved ST (John Doe) and Product (Personal loan). The generated  story result is depicted below in Figure 2: 

\begin{figure}
    \centering
    \includegraphics[width=\textwidth]{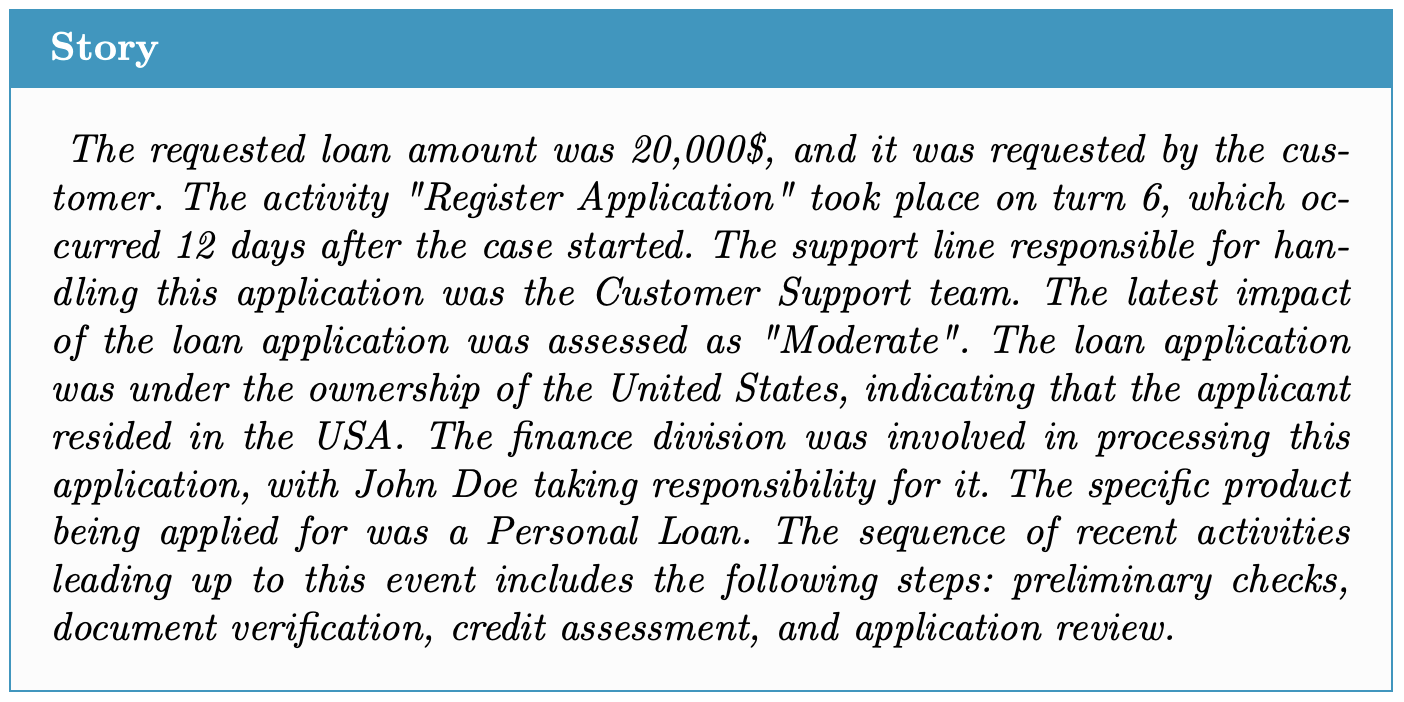}
    \caption{A generated story for a loan application example. The prompt in Figure \ref{fig:prompt-example} was used as an input to an LLM (Llama-2)}
    \label{fig:enter-label}
\end{figure}

We note that the number of tokens in the story may exceed the LFM's limit for tokens, especially when fine-tuning small models such as BERT and DeBERTa. That is amplified when the largest case length of several datasets surpass $150$ activities. To mitigate this, SNAP imposes a maximum trace length limit for the number of last activities we introduce in the story. The procedure for selection of this parameter is described in Section \ref{subsec:validation_setup}.

Another point to consider when transforming traces into stories is numerical features. LFM performs best when the story content is rich in semantic meaning, while numbers offer little such content. However, it turns out that numerical features, such as time from process start in days, can improve the model's goodness-of-fit.  

\textbf{Fine-tuning LFM} -
The transformation of the event log into stories produces a dataset containing sample-label pairs. Each sample represent a story of a prefix trace, and the label represents the ground-truth of the next activity.
With this story-based dataset at hand, we fine-tune language foundation models (e.g., BERT and DeBERTa) with their compatible tokenizers on the next activity classification task. Ultimately, in the fine-tune training mode, the model's embedding of the [CLS] token is passed through a fully connected layer followed by a ReLU activation to acquire the category classification prediction. This token is designed for classification, as it is a global token that captures the full semantic context of the input. Technically, the model outputs a list of values, one for each activity class present in the dataset. For a single activity prediction, we select the class with the largest value. However, for a recommendation system, we may want to present the best $k$ options at any given time, so we select the largest $k$ classes in the outcome (see the MIP dataset analysis in \cite{zeltyn2022mip}).   

\section{Experimental results}
\label{sec:validation}

\subsection{Datasets and benchmarks}

\label{subsec:datasets}

Table \ref{table:datasets} describes six publicly available datasets that are used for the validation of our prediction models. All datasets, except MIP, are real-world event logs that can be downloaded from the {\em 4TU Center for Research Data (\url{data.4tu.nl/info/en/})}.
The MIP synthetic dataset was presented and published in \cite{zeltyn2022mip}. 
The benchmark datasets cover a wide range of domains. The two BPI13 logs describe Volvo IT procedures for closed problems and incidents. The Env Permit is related to an environmental permit application process, while the Sepsis event logs represent the pathway of patients with sepsis through a hospital. The NASA dataset was obtained through the code instrumentation tools and describes events and exceptions in a software system. Finally, the MIP (Management Incentive Program) dataset has been simulated to resemble a real-world HR use-case in the conversational RPA domain.  

\begin{table}[h!]
\begin{center}
\begin{tabular}{lccccc}
\toprule
Dataset & Num.   & Num.    & Num.       & Avg.\ case     \\
        & cases  & events & activities  & length       \\
%
\midrule
BPI13cp    & 1487  & 6660   & 7   & 4.48  \\
BPI13in    & 7554  & 65533  & 13  & 8.68   \\
Env Permit & 1434  & 8577   & 27  & 5.98  \\
Sepsis     & 1049  & 15214  & 16  & 14.48  \\
Nasa       & 2566  & 73638  & 94  & 28.70  \\
MIP        & 1000  & 49604  & 36  & 49.60  \\
\bottomrule
\end{tabular} 
\end{center}
\caption{Description of benchmark datasets}
\label{table:datasets}
\end{table}

%
The selection of datasets in Table \ref{table:datasets} was strongly affected by \cite{ramamaneiro2021benchmark}, who compared nine state-of-the-art NAP algorithms. We validate the novel SNAP algorithm on several event logs analyzed in this paper. Specifically, we selected datasets that contain activity names and attributes with semantic meaning. We did not consider Helpdesk dataset and several BPI12 datasets, because these process logs have little or no semantic information that constitutes the focus of our paper. 

\subsection{Semantic stories on benchmark datasets}
\label{subsec:semantic_stories}
%
%
Our method for the design of semantic stories is presented in Section \ref{subsec:snap}.  
We use the XGBoost classification model for the feature selection step. We apply XGBoost for the NAP task using all available features and enriching the feature space with a frequency-based encoding methodology \cite{ramamaneiro2021benchmark}. We then select up to six features with the importance score above a set threshold. XGBoost cannot incorporate free-text attributes with practically limitless possible values (e.g., user utterances in MIP). We therefore automatically added them to the selected features list.

After the feature selection stage, the Llama-2 language model \cite{llama2} was applied to generate semantic stories with the prompt described in Section \ref{subsec:snap}.
 
Here we illustrate SNAP story design on the MIP dataset. In the MIP dataset \cite{zeltyn2022mip}, the user, a manager of a software engineering team, engages with the system to view different performance reports and, subsequently, to increase a salary of employees with the best performance.

\begin{table}[h!]
\begin{center}
\begin{tabular}{p{\linewidth}}
\toprule
Dataset  Example    \\
%
\midrule
\textbf{Features in the event log:} session number, role, user id, timestamp, activity, turn, user utterance, chatbot response, intent, intent confidence, entity, entity confidence, score, expecting response
 \\
\textbf{Selected features:} role, turn number, session number, user utterance, chatbot response, sequence of skills \\

\textbf{Semantic story example:} During the third session, a \textit{team leader} initiated a request, marking the fourth \textit{turn} of the conversation. The \textit{request} was to "view project assessment report", reflecting the user's intent to scrutinize project-related performances. The chatbot, in \textit{response}, presented the "project assessment report", aligning its reply with the user's demand. This interaction is part of a broader \textit{conversational flow} that began with a welcoming message, proceeded to report yearly assessments, engaged in disambiguation to clarify queries, and culminated in reporting project assessments. This sequence showcases the chatbot's adeptness in navigating through a sequence of skills, tailored to address the evolving needs of its users in the context of HR candidate promotion.  \\
\textbf{Trace label:}  {\em Report learning activities} \\
\bottomrule
\end{tabular} 
\end{center}
\caption{MIP dataset: Illustration to story design pipeline}
\label{table:stories}
\end{table}

Table \ref{table:stories} explains the pipeline of the story design. We select features from the log and generate the story template using LLM. Event log data is then transformed using the story template to acquire the processed dataset that consists of pairs of textual semantic stories and their labels (actual next activities). We then train the fine-tune phase of the pre-trained LFM for a classification task, as described in Section \ref{subsec:snap}.

\subsection{Fine-tuning setup and implementation}

\label{subsec:validation_setup}
%
%
In our fine-tuning SNAP experiments with bi-directional language models, we run two prediction algorithms based on BERT and DeBERTa, respectively. The experiment setup is consistent with the benchmark paper \cite{ramamaneiro2021benchmark}. We used 5-fold cross-validation with a 64-16-20 train-validation-test split. For each trace, predictions start with the second activity of the process instance and continue up to the trace end. Finally, an "end of trace" activity is added to the set of activities with the label "end".   

For each algorithm and dataset, we computed the average accuracy and F1 score that were weighted by the number of cases in each class. Both metrics were also used in \cite{ramamaneiro2021benchmark} and \cite{ramamaneiro2023graph}.
Dropout rate was equal to 0.5 in all experiments. Learning rate and batch size were optimized on the validation sets. Maximum window backward trace length constituted an additional hyper-parameter and was also optimized on validation sets. Ultimately, most frequently used hyper-parameters were batches of size 4, learning rate of $10^{-5}$, and a window size equal to 10 past activities. This combination of parameters was used for 4 datasets out of 6. A window of 15 activities has been optimal for Sepsis, and a batch of size 8 was used for MIP. A maximum number of 15 epochs were run for each experiment. The training stopped beforehand if validation accuracy did not increase for three consecutive epochs. The model with the best validation accuracy was saved and validated on the testing set. The algorithms with bi-directional LM were implemented using the Pytorch and Transformers packages and run on x86 compute nodes with Nvidia V100 GPU and an 80GB memory requirement.  Experiments with GPT-3 fine-tuning were performed using the OpenAI API and Babbage model.


\subsection{Results}

\label{subsec:validation_results}

\paragraph{Comparing SNAP and the state of the art.} 
In our main validation experiment, we fine-tuned several well-known LFMs and compared their performance with the benchmarks. First, we considered two bi-directional LFMs, BERT - \textit{bert-base-cased} and DeBERTa - \textit{microsoft/deberta-base}, and named their implementations SNAP-B and SNAP-D, respectively. Then we fine-tuned OpenAI GPT-3 model and named it SNAP-G. GPT-3 model is considered much more powerful than BERT and its modifications, so we tested the performance differences between SNAP-G and the other two models. Tables \ref{table:accuracy} and \ref{table:F1_score} summarize the results of the experiment for the accuracy and the weighted F1-score, respectively. 
\begin{table}[h!]
\begin{center}
\begin{tabular}{l|cccccc}
\toprule
   & BPI13cp &  BPI13in  & Env Permit  & Sepsis  &  \ \ Nasa \ \ & MIP  \\
%
%
\midrule
Camargo, \cite{camargo2019lstm}   & 0.547  & 0.667 & 0.858  &  &  & \\
Evermann,\cite{evermann2017DL} & 0.588  & 0.668 & 0.762  & 0.400 & 0.204 & \\
Hinnka, \cite{hinnka2020rnn} & 0.635  & 0.747 & 0.844  & 0.635 & 0.885 & \\
Khan, \cite{khan2018memory} & 0.436  & 0.519 & 0.836  & 0.210 & 0.127 & \\
Mauro, \cite{mauro2019cnn} & 0.249  & 0.367 & 0.536  & 0.615 & 0.210 & \\
Pasquadibisceglie, \cite{Pasquadibisceglie2019cnn} \ & 0.475 & 0.460  & 0.867 & 0.562 & 0.883 & \\
Tax, \cite{tax2017lstm} & 0.640  & 0.701 & 0.857  & 0.642\cellcolor{yellow!50} & 0.894\cellcolor{yellow!50} &  \\
Theis, \cite{theis2019decay}     & 0.595  & 0.594 & 0.863  & 0.557 & 0.890 &  \\
Venugopal, \cite{venugopal2021comparison} & 0.484  & 0.496 & 0.696  & & &  \\
Rama-Maneiro, \cite{ramamaneiro2023graph}  & 0.675 & \bf{0.777}\cellcolor{cyan!50} & 0.877\cellcolor{yellow!50} &  &  &  \\
Zeltyn, \cite{zeltyn2022mip} & & & & & & \ 0.390 \ \\
\midrule\midrule
SNAP-B        & 0.679$^*$\cellcolor{orange!50} & 0.758 & 0.880\cellcolor{orange!50} & 0.649$^*$\cellcolor{orange!50} & \bf{0.898}$^*$\cellcolor{cyan!50} & 0.424\cellcolor{yellow!50}$^*$ \\
SNAP-D        & 0.679$^*$\cellcolor{orange!50} & 0.764\cellcolor{yellow!50} & 0.873 & 0.567 & 0.846 & 0.437$^*$\cellcolor{orange!50} \\
SNAP-G        & \bf{0.696}$^*$\cellcolor{cyan!50} & 0.774\cellcolor{orange!50} & \bf{0.890}$^*$\cellcolor{cyan!50} & \bf{0.655}$^*$\cellcolor{cyan!50} & 0.895\cellcolor{orange!50} & \bf{0.459}$^*$\cellcolor{cyan!50} \\
\end{tabular} 
\end{center}
\caption{Accuracy comparison between the SNAP-B, SNAP-D, SNAP-G, and SOTA benchmark algorithms. The best, second-best, and third-best approaches are highlighted in cyan, orange, and yellow, respectively. $^*$The SNAP models significantly outperform the SOTA with $p_{value}<0.05$ (Wilcoxon signed-rank test).}
\label{table:accuracy}
\end{table} 

\begin{table}[h!]
\begin{center}
\begin{tabular}{l|cccccc}
\toprule
   & BPI13cp &  BPI13in  & Env Permit  & Sepsis  &  \ \ Nasa \ \ &  MIP  \\
%
%
\midrule
Camargo, \cite{camargo2019lstm}  & 0.523  & 0.614 & 0.851  &  &  & \\
Evermann, \cite{evermann2017DL}  & 0.505  & 0.585 & 0.739  & 0.303 & 0.123 & \\
Hinnka, \cite{hinnka2020rnn}    & 0.571  & 0.730 & 0.825  & 0.612 & 0.877 & \\
Khan, \cite{khan2018memory} & 0.369  & 0.434 & 0.822  & 0.176 & 0.106 & \\
Mauro, \cite{mauro2019cnn} & 0.100  & 0.362 & 0.524  & 0.602 & 0.169 & \\
Pasquadibisceglie, \cite{Pasquadibisceglie2019cnn} \ & 0.427 & 0.374  & 0.849 & 0.536 & 0.874 & \\
Tax, \cite{tax2017lstm} & 0.590  & 0.684 & 0.842  & 0.633\cellcolor{yellow!50} & 0.886\cellcolor{yellow!50} &  \\
Theis, \cite{theis2019decay}     & 0.534  & 0.547 & 0.845  & 0.526 & 0.880 &  \\
Zeltyn, \cite{zeltyn2022mip} & & & & & & \ 0.330 \ \\
\midrule\midrule
SNAP-B        & 0.670$^*$\cellcolor{orange!50} & 0.754$^*$\cellcolor{yellow!50} & 0.866$^*$\cellcolor{orange!50} & 0.638$^*$\cellcolor{orange!50} & \bf{0.890}$^*$\cellcolor{cyan!50} & 0.421$^*$\cellcolor{yellow!50}  \\
SNAP-D        & 0.664$^*$\cellcolor{yellow!50} & 0.757$^*$\cellcolor{orange!50} & 0.860$^*$\cellcolor{yellow!50} & 0.540 & 0.832 \ & 0.428$^*$\cellcolor{orange!50} \\
SNAP-G        & \bf{0.678}$^*$\cellcolor{cyan!50} & \bf{0.767}$^*$\cellcolor{cyan!50} & \bf{0.876}$^*$\cellcolor{cyan!50} & \bf{0.641}$^*$\cellcolor{cyan!50} & 0.887\cellcolor{orange!50} & \bf{0.447}$^*$\cellcolor{cyan!50}  \\
\end{tabular} 
\end{center}
\caption{Weighted F1-score comparison between the SNAP-B, SNAP-D, SNAP-G, and SOTA benchmark algorithms. The best, second-best, and third-best approaches are highlighted in cyan, orange, and yellow, respectively.  $^*$The SNAP models significantly outperform the SOTA with $p_{value}<0.05$ (Wilcoxon signed-rank test).}
\label{table:F1_score}
\end{table} 
The state-of-the-art performance is taken from \cite{ramamaneiro2021benchmark,ramamaneiro2023graph}. Note that \cite{ramamaneiro2023graph} does not contain results for F1-score and does not analyze the NASA and Sepsis datasets. The MIP benchmark is taken from \cite{zeltyn2022mip} and is based on XGBoost with the frequency-based encoding methodology.

We performed the non-parametric Wilcoxon signed-rank test to check the statistical significance of the results. We separately tested each SNAP model against the SOTA. In terms of F1-score (Table \ref{table:F1_score}), SNAP significantly outperforms the SOTA on all datasets.
Our empirical analysis reveals that SNAP-G accuracy yields superior outcomes over most of the datasets (Table \ref{table:accuracy}). SNAP-B is slightly better than SNAP-G for NASA and \cite{ramamaneiro2023graph} is slightly better for BPI13in. In Table \ref{table:F1_score}, we can see that SNAP-G's F1-score outperforms other methods for all datasets, except NASA. 

As SNAP-G is a super-large model and the stories contain all the process information, it seems plausible that its results are close to the maximal achievable accuracy. Comparing SNAP-G to SNAP-B and SNAP-D, we observe that even the smaller SNAP models demonstrate solid performance. Comparing SNAP-B and SNAP-D shows that in instances where SNAP-D demonstrates higher performance than SNAP-B, the discernible advantage is of modest proportion. Conversely, in scenarios where SNAP-B outperforms SNAP-D, such as in the case of the NASA and Sepsis datasets, the observed performance difference is notably substantial.

The relatively low accuracy numbers for the MIP dataset and all prediction models are due to the large number of activities and the significant variance in the sequences it contains. Yet, SNAP significantly outperforms the benchmark models due to a large amount of semantic information in MIP. Furthermore, we see that the average enhancement margin of SNAP models is more substantial for the F1-score compared to accuracy. Therefore, SNAP seems to be more balanced between recall and precision than the state-of-the-art algorithms.

\paragraph{Does semantic story structure matter?}
Design of coherent and grammatically correct semantic stories from business process logs constitutes a key step in the SNAP algorithm. It is natural to ask if this step is necessary. One can simply concatenate feature values, including the historical sequence of activities, into a text string and use it a "basic story" input to the SNAP algorithm.    
\begin{table}[h!]
\begin{center}
\begin{tabular}{lcccc}
\toprule
Dataset &  Story &  \ \ \ List of values \ \ \ \\ 
      & acc. \ \ \ F1  &   acc. \ \ \ F1   \\
%
%
\midrule
BPI13cp     & 0.679 \ \ \ 0.670  & 0.616 \ \ \ 0.614  \\
BPI13in     & 0.758 \ \ \ 0.754  & 0.733 \ \ \ 0.731  \\
Env Permit \ & 0.880 \ \ \ 0.866  & 0.859 \ \ \ 0.845 \\
Sepsis      & 0.649 \ \ \ 0.638  & 0.635 \ \ \ 0.630 \\
Nasa        & 0.896 \ \ \ 0.890  & 0.880 \ \ \ 0.877 \\
MIP         & 0.424 \ \ \ 0.421  & 0.426 \ \ \ 0.421 \\
\bottomrule
\end{tabular} 
\end{center}
\caption{SNAP-B comparison: Semantic stories versus lists of feature values.} 
\label{table:features_comparison}
\end{table}

Table \ref{table:features_comparison} compares the performance metrics for the two approaches using the same 5-fold fine-tuning setup. We observe that the semantic representation significantly improves the accuracy and F1-score of our approach for all datasets except MIP, where they are comparable. I fact, the semantic information in MIP is contained mainly in activity names and text utterances; coherent story seems lees important for MIP.
\paragraph{Does semantic information in activity names matter?}
In SNAP, we gave meaningful names to the activities and performed some preprocessing work to replace shortened or unclear activity names in the logs. We tested the model's performance when the activity names were replaced with mere numbers in the stories. We mapped each activity name to an integer and replaced them, thus removing its semantic content. 
\begin{table}[h!]
\begin{center}
\begin{tabular}{lccc}
\toprule
Dataset & Named activities  & \ \ Numbered activities \ \  \\
        & acc. \ \ \ \ F1   &  acc. \ \ \ \ F1   \\        
%
\midrule
BPI13cp     & 0.679 \ \ \ 0.670  & 0.670 \ \ \ 0.661  \\
BPI13in     & 0.758 \ \ \ 0.754  & 0.758 \ \ \ 0.753  \\
Env Permit  & 0.880 \ \ \ 0.866  & 0.877 \ \ \ 0.865 \\
Sepsis      & 0.649 \ \ \ 0.638  & 0.648 \ \ \ 0.640 \\
Nasa        & 0.896 \ \ \ 0.890  & 0.896 \ \ \ 0.889 \\
MIP         & 0.424 \ \ \ 0.421  & 0.401 \ \ \ 0.395 \\
\bottomrule
\end{tabular} 
\end{center}
\caption{SNAP-B comparison: Semantic story with activity names versus stories with numbered activities}
\label{table:number_activities}
\end{table}

Table \ref{table:number_activities} compares SNAP-B for the two cases. We observe a significant performance decline for MIP, a moderate decline for BPI13cp and Env Permit and non-significant decline for other datasets. We conclude that the importance of semantically meaningful activity names strongly depends on the dataset.

\paragraph{Do user utterances in conversational RPA system matter?}
In contrast to classical business processes, the logs of conversational RPA systems contain rich textual information of the user utterances and bot responses. We checked if this information is vital for prediction quality and ran SNAP algorithm with BERT and DeBERTa models excluding user utterances on MIP dataset. The best results were achieved via DeBERTa with accuracy of 0.373 and weighted F1-score of 0.358, significantly inferior to 0.437 and 0.428, respectively, in Tables \ref{table:accuracy} and \ref{table:F1_score}.
These experiments, jointly with those summarized in Tables \ref{table:features_comparison} and \ref{table:number_activities}, indicate that various types of semantic information within the event logs pose meaningful value for the LFM-based predictions. They also demonstrate that SNAP can be especially relevant for logs with rich textual and conversational data.   
 
\section{Discussion and Limitations}

\label{sec:discussion}

We introduce the SNAP model, a novel approach based on constructing semantic stories from event logs. The core concept of semantic stories presents a highly efficient methodology for incorporating activities and their valuable event-log attributes into a cohesive narrative. 
That is particularly useful where the categorical feature space is huge, such as user utterances and other free-text attributes. Moreover, the usage of BERT-like models with the stories results in a semantic bi-directorial encoding of information of the trace.
Through comprehensive evaluation on six benchmark datasets we demonstrate that SNAP outperforms existing state-of-the-art models.

\textit{Semantic information:}
A key limitation of our model lies in the level of semantic information available in the event log (both in the attributes name and values). We conjecture that the SNAP model will not improve the SOTA benchmark in cases with limited semantic information, such as event logs where all attributes are numbers or merely two to three features are available. In such cases, we recommend using the more traditional ML techniques (e.g., XGBoost and CatBoost). 
A central finding in this work is that the performance of SNAP increases with the amount of semantic information within the dataset. That is particularly relevant in the context of conversational RPA systems and digital assistants, where the user and system utterances often contain rich semantic information that can be used to improve the accuracy of predictions.

\textit{Modest improvements:}
While our proposed model demonstrates improvement over the SOTA, it is important to acknowledge that the observed improvements, though statistically significant, might appear modest when evaluated from a practical perspective. The average enhancement achieved by SNAP amounts to approximately 7\%, which might not immediately translate into substantial real-world impact. 
The observed improvement is partly attributed to the inherent challenges within the datasets. The NAP task in the BPM domain is firmly established, with decades of research and modeling contributing to the current SOTA. Achieving substantial advancements in such a mature field is inherently difficult, especially when dealing with datasets with minimal semantic information. We believe that on semantically rich real-world datasets (e.g., in the conversational RPA domain), SNAP's advantage will increase. In addition, it is worth noting that the experiments conducted using GPT-3 also introduce a grasp of the approximation for the upper-bound accuracy of this benchmark. As GPT-3 is two orders of magnitude larger than the SOTA and SNAP-B models and we included all the available attributes, we conjecture that SNAP-G results are close to the optimal results. 

\textit{Data security:}
In some cases, customers will prefer not to use LLMs such as ChatGPT and Llama 2. That might happen when the features' names are sensitive or when the data is confidential. In such cases, we suggest to manually construct the story template, as this step is carried out only once per event log. We found out that in many use cases, the features are relatively similar and typically contain such items as identity, time, role, activity, turn, and a couple of additional attributes. In conversational RPA systems, user utterance and system response are also included. Thus, we believe that SNAP can be easily implemented in such use-cases with little manual work.

\textbf{Future research:} 
The main idea of the SNAP algorithm can easily fit other predictive process monitoring tasks, such as outcome prediction, remaining time prediction, and future process cost. We aim to explore SNAP's performance in these types of PBPM tasks. By adjusting the story style and the LFM's head layer, SNAP can be adapted to other predictive tasks.
In addition, it is possible to extend the LFM's pre-training phase with an appropriate new masking task. Such an extension has the potential to boost SNAP's performance in new tasks.
 
\bibliographystyle{splncs04}
\bibliography{bib}

\end{document}